\documentclass[acmtog]{acmart}
\acmSubmissionID{1545}

\usepackage{booktabs} % For formal tables
\usepackage{pgfplots}
\usepackage{colortbl}
\usepackage{hyperref}
\usepackage[ruled]{algorithm2e} % For algorithms

\SetAlFnt{\small}
\SetAlCapFnt{\small}
\SetAlCapNameFnt{\small}
\SetAlCapHSkip{0pt}

\copyrightyear{2025}
\acmYear{2025}
\setcopyright{cc}
\setcctype{by}
\acmConference[SA Conference Papers '25]{SIGGRAPH Asia 2025 Conference Papers}{December 15--18, 2025}{Hong Kong, Hong Kong}
\acmBooktitle{SIGGRAPH Asia 2025 Conference Papers (SA Conference Papers '25), December 15--18, 2025, Hong Kong, Hong Kong}\acmDOI{10.1145/3757377.3763892}
\acmISBN{979-8-4007-2137-3/2025/12}

\begin{document}
% Title portion
\title{GS-RoadPatching: \\ Inpainting Gaussians via 3D Searching and Placing for Driving Scenes}

% DO NOT ENTER AUTHOR INFORMATION FOR ANONYMOUS TECHNICAL PAPER SUBMISSIONS TO SIGGRAPH 2019!
\author{Guo Chen}
\authornote{Guo Chen and Jiarun Liu have equally contributed to this work. \\ $\dag$ Sheng Yang and Shi-Sheng Huang are joint corresponding authors.}
\orcid{0009-0008-1528-2262}
\affiliation{%
 \institution{Beijing Normal University}
 \country{China}}

\author{Jiarun Liu}
\authornotemark[1]
\orcid{0009-0003-2835-4708}
\affiliation{%
 \institution{Unmanned Vehicle Dept., Cainiao, Alibaba \& State Key Lab of CAD\&CG, Zhejiang University}
 \country{China}}

\author{Sicong Du}
\orcid{0000-0001-7942-5068}
\affiliation{%
 \institution{Unmanned Vehicle Dept., Cainiao, Alibaba}
 \country{China}}

\author{Chenming Wu}
\orcid{0000-0001-8012-1547}
\affiliation{%
 \institution{Baidu}
 \country{China}}

\author{Deqi Li}
\orcid{0000-0001-5742-6609}
\affiliation{%
 \institution{Beijing Normal University}
 \country{China}}

\author{Shi-Sheng Huang}
\authornotemark[2]
\orcid{0000-0002-6978-8022}
\affiliation{%
 \institution{Beijing Normal University}
 \country{China}}
\email{huangss@bnu.edu.cn}

\author{Guofeng Zhang}
\orcid{0000-0001-5661-8430}
\affiliation{%
 \institution{State Key Lab of CAD\&CG, Zhejiang University}
 \country{China}}

\author{Sheng Yang}
\authornotemark[2]
\orcid{0000-0001-8295-1552}
\affiliation{%
 \institution{Unmanned Vehicle Dept., Cainiao, Alibaba}
 \country{China}}
\email{shengyang93fs@gmail.com}

\renewcommand\shortauthors{Chen, G. and Liu, J. et al}
\renewcommand\shorttitle{GS-RoadPatching: Inpainting Gaussians via 3D Searching and Placing for Driving Scenes}

\begin{abstract}

This paper presents GS-RoadPatching, an inpainting method for driving scene completion by referring to completely reconstructed regions, which are represented by 3D Gaussian Splatting (3DGS). Unlike existing 3DGS inpainting methods that perform generative completion relying on 2D perspective-view-based diffusion or GAN models to predict limited appearance or depth cues for missing regions, our approach enables substitutional scene inpainting and editing directly through the 3DGS modality, extricating it from requiring spatial-temporal consistency of 2D cross-modals and eliminating the need for time-intensive retraining of Gaussians. Our key insight is that the highly repetitive patterns in driving scenes often share multi-modal similarities within the implicit 3DGS feature space and are particularly suitable for structural matching to enable effective 3DGS-based substitutional inpainting. Practically, we construct feature-embedded 3DGS scenes to incorporate a patch measurement method for abstracting local context at different scales and, subsequently, propose a structural search method to find candidate patches in 3D space effectively. Finally, we propose a simple yet effective substitution-and-fusion optimization for better visual harmony. We conduct extensive experiments on multiple publicly available datasets to demonstrate the effectiveness and efficiency of our proposed method in driving scenes, and the results validate that our method achieves state-of-the-art performance compared to the baseline methods in terms of both quality and interoperability. Additional experiments in general scenes also demonstrate the applicability of the proposed 3D inpainting strategy. The project page and code are available at: 
\href{https://shanzhaguoo.github.io/GS-RoadPatching/}{\color{blue}https://shanzhaguoo.github.io/GS-RoadPatching/}.

\end{abstract}

%
% The code below should be generated by the tool at
% http://dl.acm.org/ccs.cfm
% Please copy and paste the code instead of the example below.
%


%
% End generated code
%


\begin{teaserfigure}
\includegraphics[width=\textwidth]{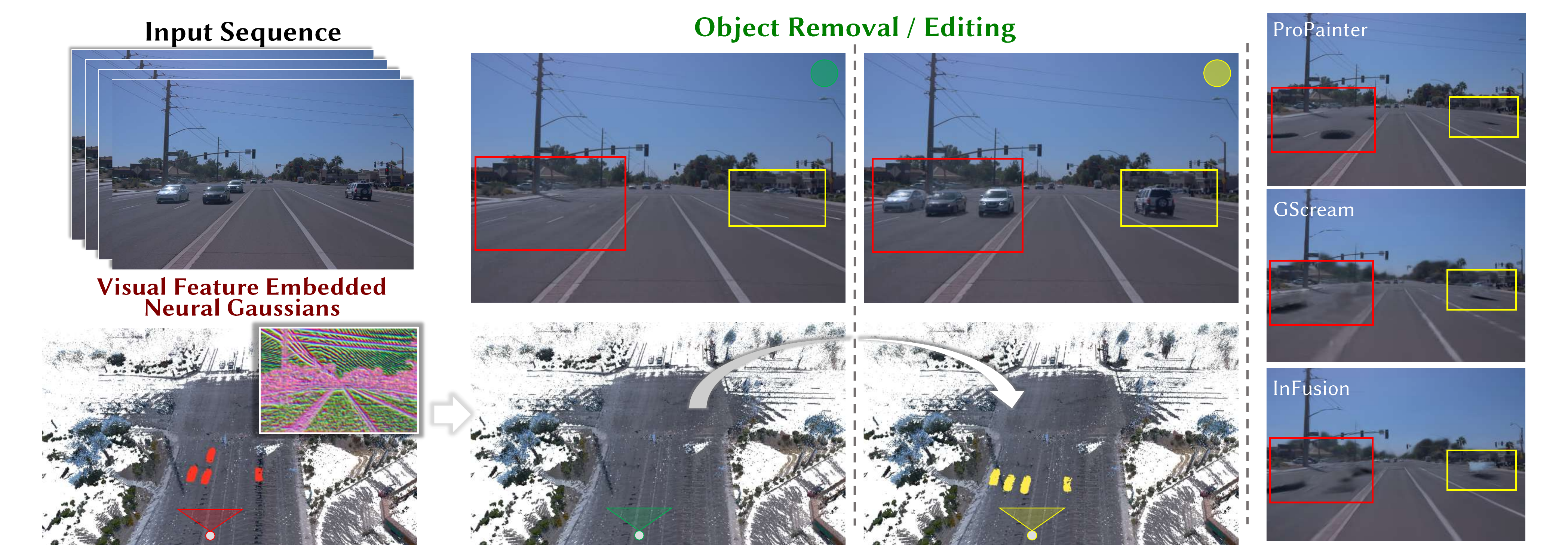}
\caption{
GS-RoadPatching reconstructs street scenes from raw driving sequences by 3D Gaussian Splatting (3DGS) and provides applications—moveable object removal and target editing—by directly manipulating 3DGS assets based on our proposed patch-based inpainting strategy. Compared with previous 2D video inpainting approaches (e.g., ProPainter~\cite{zhou2023propainter}) and 3DGS inpainting approaches (e.g., InFusion~\cite{liu2024infusion} and GScream~\cite{wang2025gscream}), our GS-RoadPatching can achieve more complete and harmonic scene inpainting for more realistic object removal and editing, especially for driving scenes.
% {\chenming{Replace it with a teaser image.} \sh{Prepare layout and detailed results. @CG, JR, Chenming.} \sh{First edition of teaser caption. @Sheng.}}
}
\label{fig:teaser}
\end{teaserfigure}

\maketitle

\section{Introduction}
\label{sec:intro}

Modeling a virtual environment that supports photorealistic simulation~\cite{hu2023uniad} and physically accurate interaction~\cite{nvidia2025cosmos} through 3D reconstruction and rendering~\cite{wu2024survey} is crucial for autonomous driving. As an emerged explicit representation in recent years, 3D Gaussian Splatting (3DGS)~\cite{Kerbl20233dgs} demonstrates promising performance in terms of real-time rendering, and enables scene editing applications compared to neural radiance field (NeRF)~\cite{mildenhall2021nerf}. Building on 3DGS, various methods have been proposed to enhance flexibility and improve scene editing quality~\cite{bi2024gs3}. These include techniques like generative completion~\cite{fan2024freesim} and substitutional inpainting~\cite{liu2024infusion}, among others. While generative completion enables creative content generation, \textit{substitutional inpainting} ensures interpretable and controllable editing processes. For highly structured and repetitive road scenes, replacing incomplete regions with well-reconstructed counterparts from other areas presents a particularly practical solution.

% 3D Reconstruction and rendering~\cite{wu2024survey} create physical-realistic content~\cite{nvidia2025cosmos} and closed-loop simulation for autonomous driving~\cite{hu2023uniad}. 3D Gaussians~\cite{Kerbl20233dgs}, as an emerged explicit representation in recent years, have significantly improved the real-time rendering quality, and enabled the practicality of 3D scene editing. While a variety of reconstruction methods~\cite{kerbl2024h3dgs,ren2024octree} gradually improve their quality and volume, reasonably editing scenes including substitution~\cite{liu2024infusion} and generative completion~\cite{fan2024freesim} also matters for downstream applications~\cite{}.

To achieve reasonable substitutional inpainting, most existing methods~\cite{liu2024infusion,mirzaei2024reffusion,wang2025gscream} adopt a 2D prior-guided approach: first, performing 2D inpainting to obtain depth or diffusion priors, followed by a second round of reconstruction~\cite{yu2024gaussian}. In contrast, shape analysis methods for directly manipulating 3D Gaussians are still under-explored. Inspired by techniques developed for 3D point clouds and 3D meshes~\cite{wang20243dsurvey}, we advocate for conducting structural analysis and scene editing on 3D Gaussians. This approach offers two key advantages with sustainable geometry and photo-realistic rendering quality: (1) enabling direct operations on compact 3D Gaussian scenes instead of re-generating raw training data for additional rounds of reconstruction, and thus (2) eliminating the dependence on the spatial-temporal consistency of 2D video diffusion models~\cite{yan2024streetcrafter,xu2024streetunveiler}, which is typically necessary to ensure high-quality multi-view reconstruction.

Our core idea borrows from PatchMatch~\cite{barnes2009Patchmatch,wang2021patchmatchnet}, which has been widely used for 2D image inpainting based on structural analysis. Inspired by this patch-based searching-measuring-replacing strategy, we focus on conducting such a scheme on street scenes represented by 3D Gaussians.
This involves four core challenges to be addressed by our work: (1) Automatically locating incompletely reconstructed regions or manually identifying undesired objects. (2) Evaluating the compatibility between each target region and candidate sources, selecting the best match based on inherent similarity and contextual consistency. (3) Replacing target regions with selected candidate sources and applying patch fusion for better visual results.
% \dubious{Specifically, we use a spatial octree structure for scene organization and construct patches through voxel partitioning.}
(4) Scaling such a patch-based strategy at different sizes to balance global structural consistency and local detailed appearance.
% \dubious{We also embed the 2D visual feature into structured 3DGS anchors to better locate and search for patches with visual similarity. Finally, we design a simple yet effective fusion method for better visual harmony, which only involves several iterations to finetune the reconstructed scene.}
% \sh{These two sentences are either separated strategies from our motivation, or already mentioned in (3)? Maybe we can omit them here and mention in Sec. 3?}

To tackle these challenges, we produce several patch-based operators -- (1) Locating and segmenting incomplete regions, (2) Searching and measuring with optimal candidates, and (3) pasting and fusing to the target region. Notably, these operators are further (4) scalable through patch-anchor indexing for editing regions. Prominently, GS-RoadPatching provides the following strategies as main contributions:
\begin{itemize}
    \item A novel inpainting pipeline based on the 3DGS for driving scenes, introducing patch-based operators for high-fidelity scene completion, which results in superior performance compared to existing 2D-prior guided inpainting methods.
    \item A visual feature embedded 3DGS framework that integrates deep visual features into Gaussian attributes, significantly improving patch similarity measurement and retrieval accuracy in street scenes compared to conventional geometry and appearance-based approaches.
    \item A simple yet effective fusion approach that harmoniously integrates reprojection consistency supervision, achieving balanced preservation of both geometric accuracy and visual coherence.
    % \item We conduct extensive experiments on driving scenes and general scenes to demonstrate the effectiveness of our proposed method, and is superior to existing 2D-prior guided inpainting methods.
\end{itemize}

Experiments on the Waymo Open dataset~\cite{waymodataset} have shown our ability to improve the inpainting quality and efficiency of driving scenes, and we have also proven our adaptability on general clips on the 360-USID dataset~\cite{wu2025aurafusion360}.
% Ablations on available hierarchical representations and operator strategies have also consolidated the necessity of data structures and loss functions.
We will release our source code to facilitate future work.

\section{Related Work}
\label{sec:rel}

\begin{figure*}[t]
\includegraphics[width=\linewidth]{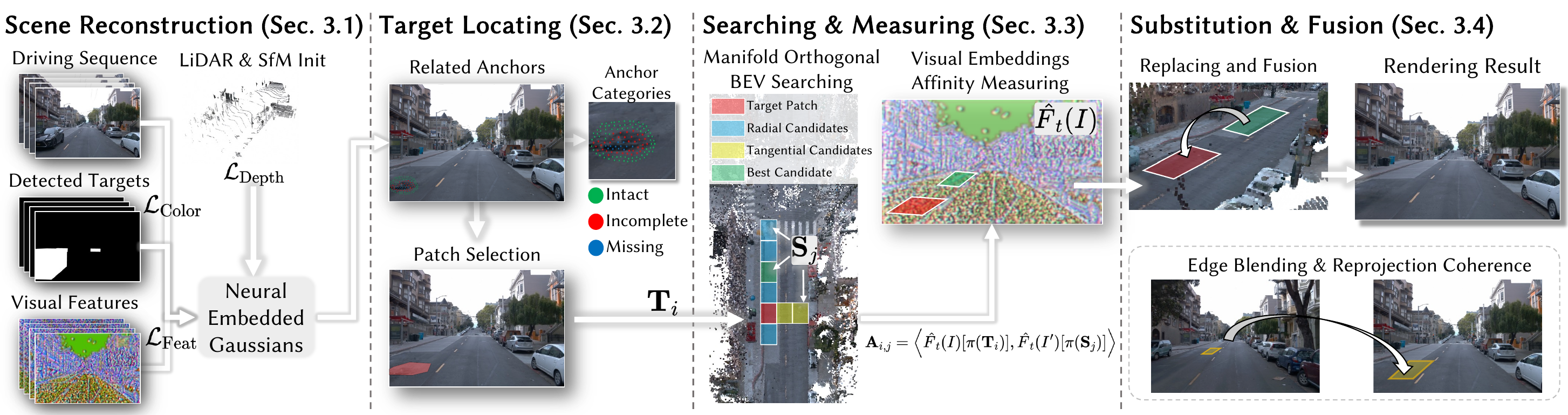}
\caption{Our GS-RoadPatching directly conducts structural analysis and scene editing on 3D Gaussians, and proposes locating, searching, substitution, and fusion for inpainting. See \textbf{method overview} in Sec.~\ref{sec:method} for the pipeline.}
\label{fig:pipeline}
\end{figure*}

\subsection{Revisiting inpainting methods of Neural Scenes}

Unlike \emph{directly editing on 2D sequences} without 3D structure~\cite{zhou2023propainter, bian2025videopainter}, \emph{diffusion-aided reconstruction} methods~\cite{fan2024freesim,ni2024recondreamer,zhuang2023dreameditor} or \emph{controllable driving world models}~\cite{yan2024streetcrafter,Zhao2024drivedreamer4d} proposed for generative completion tasks, \emph{neural scenes inpainting} -- using reconstructed areas to perform reasonable editing -- mainly refers to existing scene contexts. There are two categories of inpainting methods for neural scene representations.

The first category generates 2D priors and executes an additional round of scene re-training. Early NeRF-based~\cite{mildenhall2021nerf} methods like SPIn-NeRF~\cite{mirzaei2023spin}, NeRF-In~\cite{shen2023nerfin}, and NeRFiller~\cite{weber2024nerfiller}, to varying degrees, introduce RGB or depth priors from 2D generative methods as guidance and supervision to retrain or finetune the scene. In recent years, based on Gaussian Splatting representation~\cite{Kerbl20233dgs}, GaussianEditor~\cite{chen2024gaussianeditor} first proposes to use inpainting techniques for removing user-specified Gaussian areas and fill them adaptively. GaussianPro~\cite{cheng2024gaussianpro} introduces the propagation idea of patch-match, which propagates depth and normal information through adjacent pixels and controls Gaussian initialization to obtain better geometric and appearance representations during construction. InFusion~\cite{liu2024infusion} combines depth completion and color inpainting to achieve better geometric consistency. RefFusion~\cite{mirzaei2024reffusion} introduces score distillation loss and trains a diffusion model for the target area on the given reference view. RI3D~\cite{paliwal2025ri3d} leverage 
2D diffusion prior to perform few-shot 3DGS inpainting. StreetUnveiler~\cite{xu2024streetunveiler} introduces semantic labeled 2D Gaussians~\cite{huang20242d} for object removal while ensuring consistency through continuous reference based inpainting.

The latter category leverages multi-view 2D observations or latent embeddings for categorizing and optimizing Gaussian primitives. Feature-3DGS and Feature Splatting tend to embed semantic features into Gaussian primitives and remove the target Gaussians directly. Similarly, Gaussian Grouping~\cite{ye2025grouping} come up with semantic segmentation and editing for 3D Gaussians. DecoupledGaussians~\cite{wang2025decoupledgaussian} leverages Poisson fields for better physics-aware object decoupling in 3DGS scenes. Due to the lack of spatial constraint and detailed supplement, these methods often generate large artifacts. More recently, GScream~\cite{wang2025gscream} effectively utilizes the Scaffold-GS~\cite{lu2024scaffold} concept and employs cross-attention between the target anchor and its surrounding anchors, guided by a 2D inpainted prior, to generate a more harmonious editing area. A more recent work AuraFusion360~\cite{wu2025aurafusion360} incorporates multi-view and depth information for better multi-view consistency.

While both of these categories leverage 2D observations for 3D Gaussian re-training and optimization, they heavily rely on multi-view reprojection consistency across video frames -- an \emph{over-challenging} problem -- to fight against artifacts: (1) Inpainted 2D regions should be physically coherent for clarified 3D reconstruction, and (2) Contextual harmonious for the surrounding of the inpainted 3D target are ensured by frame-level blending. We found that for the regions to be inpainted in neural scenes, as long as there is a small amount of supervision from the original perspective images, the existing information in the 3D scene can be used for prediction and completion, thereby reducing the need to introduce additional 2D priors.

\subsection{3D Shape Analysis on Explicit Representations}

Using patch-match for inpainting visual media is an intuitive strategy, as these methods above~\cite{cheng2024gaussianpro,wang2024gaussurf} have been utilized on 2D assets. However, using such a strategy directly on 3D assets requires an appropriate data structure for traversing and effective operators for measuring. Although 3D-PatchMatch~\cite{cai20153dpatchmatch} attempted to extend the patch-match method to point clouds, this approach remains limited to shape completion tasks and exhibits poor perceptual performance on large-scale 3D assets. To the best of our knowledge, the graphics community still lacks effective methods based on 3D Gaussians so far. Hence, we briefly summarize similar strategies proposed for relevant 3D representations including scattered point clouds and hierarchical voxel grids~\cite{wang20243dsurvey}, as a guidance for our design of 3D shape analysis methods.

For completing 3D shapes and scenes represented by point cloud or Signed-Distance-Field (SDF) volume, finding similar points or patches to the missing parts -- either through traditional~\cite{cai20153dpatchmatch,huang2016structure} or learning-based approaches~\cite{mittal2021self} -- is of vital importance for reasonable editing~\cite{Tesema2024completionsurvey}. Structural similarities, symmetry, and regularity~\cite{huang2015support} are common and crucial hints for a supervised matching~\cite{mittal2021self} with the rest of shapes or templates, or an unsupervised matching~\cite{cui2022energybasedresiduallatenttransport} linking partial scans to complete scans.

Specifically during matching, as summarized by another specific survey~\cite{liu2024descriptorssurvey}: Distinctiveness, Robustness, Compactness, and Efficiency are four major concerns for local descriptors. In addition to multi-level local features, multi-resolution encoder with combined multi-layer perception~\cite{huang2020pfnet} is also an effective strategy for acquiring reliable geometric and semantic information at scales, which is suitable for hierarchical point-level features but cannot be directly applied to patch-level features in our scenario. Our proposed method directly utilizes the explicit spatial indexing and implicit embedded features from training for patch-based shape analysis, building effective similarity measurement for efficient searching and measuring.

\section{GS-RoadPatching}
\label{sec:method}

GS-RoadPatching aims to remove automatically detected or manually specified regions (e.g., movable vehicles or pedestrians) from a driving scene and substitutionally
complete the scene using the context within the scene based on the idea of 3DGS-based patch matching, while ensuring coherent and realistic novel-view synthesis for the substituted regions. %appears coherent and realistic. 

%\textbf{Method Overview.} 
Specifically, given a frame sequence of a driving scene, we first perform scene reconstruction based on neural Gaussians with efficient scene structure organization, and embedding visual features into each neural Gaussian for highly accurate Gaussian patch search and comparison (Sec.~\ref{sec:method:recon}).
Next, we locate the target patches $\mathcal{T} \triangleq \bigcup_i \{ \mathbf{T}_i \}$ that are affected by the removal, i.e., those regions should be substitutionally inpainted (Sec.~\ref{sec:method:locate}).
Subsequently, for each target patch $\mathbf{T}_i$, we perform heuristic searching in the orthogonal manifold bird eye view (BEV) space to acquire a group of candidate source patches $\mathcal{S}_i \triangleq \bigcup_j \{ \mathbf{S}_j \}$, and compare the affinity $\mathbf{A}_{i,j} \triangleq \mathbf{T}_i \ominus \mathbf{S}_j$ through their primitives containing them (Sec.~\ref{sec:method:affinity}). 
Finally, we perform patch replacement to substitute the original target patch $\mathbf{T}_i$ and perform post-fusion $\mathbf{B}_{i,j} \triangleq \mathbf{S}_j \otimes \mathbf{T}_i$ for local harmony (Sec.~\ref{sec:method:fusion}). Fig.~\ref{fig:pipeline} shows the main pipeline of our GS-RoadPatching. %in Fig.~\ref{fig:pipeline}. 
% \sh{Where is the mentioned  `spatial voxel partitioning' in the contribution? We should highlight that.}

\subsection{Scene Reconstruction with Visual Feature Embedding}
\label{sec:method:recon}

Compared to the original 3DGS, neural Gaussians~\cite{lu2024scaffold} offer significant advantages in terms of anchor-based structural efficiency and geometry-aware preservation of spatial relationships. This makes them especially suitable for urban environment reconstruction, like driving scenes. %(1) Structural efficiency -- the anchor-based hierarchy enables the compact organization of millions of Gaussians through topological abstraction, and (2) Geometric awareness -- the MLP-inferred attributes preserve spatial relationships across scales, crucial for urban environments. 
However, one major drawback for the original neural Gaussian is the inability to precisely measure the similarity between patches. For better structural similarity measurement of neural Gaussian, we propose to embed additional properties on each Gaussian with perspective-specific visual feature embeddings that serve as complementary descriptors.

\begin{figure}[t]
\centering
    \includegraphics[width=0.98\linewidth]{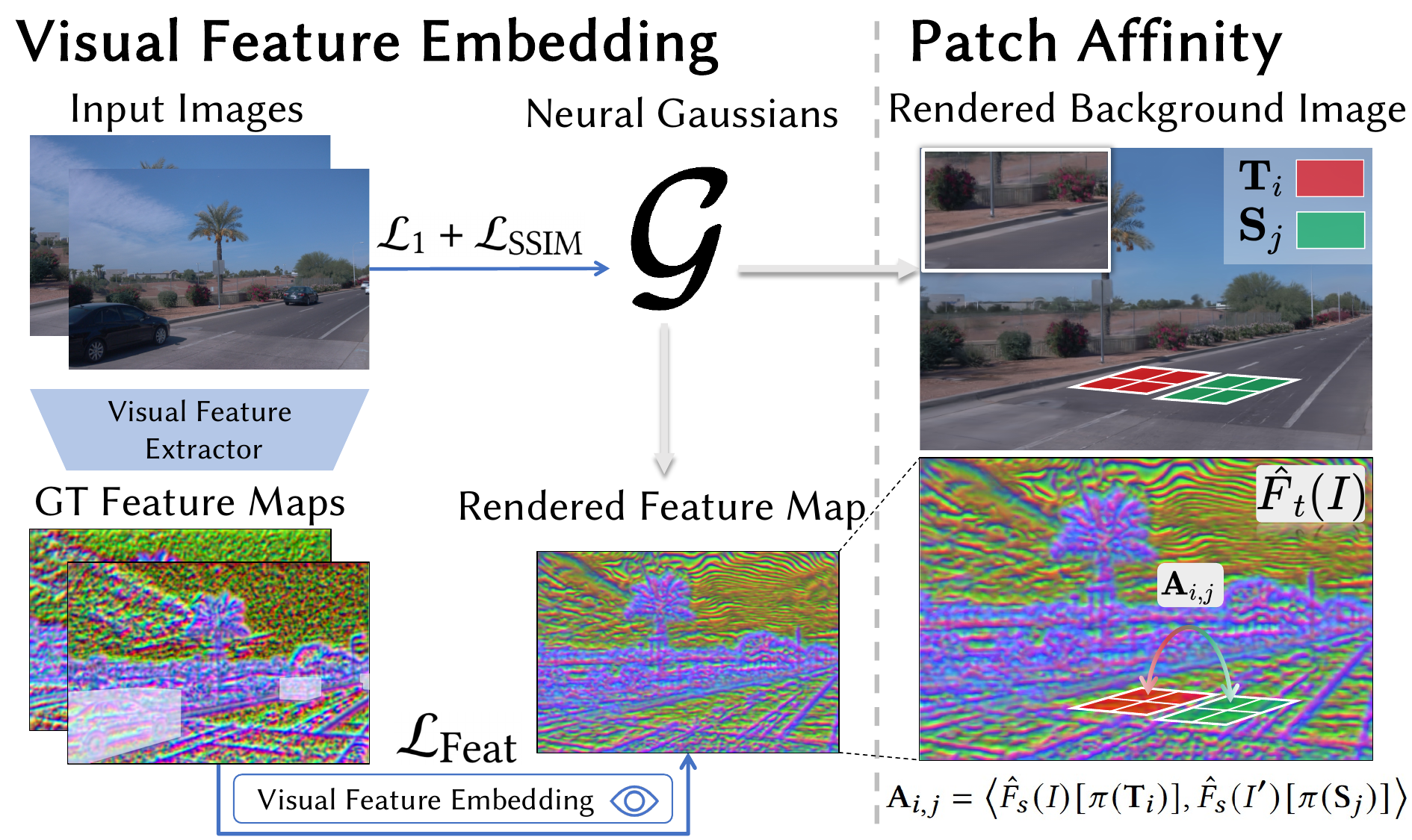}
    \caption{Illustration of visual feature embedding and patch similarity. We employ the deep visual feature map extracted from the input image as the GT feature map in the supervised training of visual feature embeddings. This approach enhances similarity computation based on 3D patch projection results, enabling optimal candidate patch selection.}
    \label{fig:feature}
\end{figure}

\textbf{Visual Feature Embedding.}
% Specifically, for each anchor $\mathbf{v}$, we learn its Gaussian feature $\mathbf{g_v}$ and visual embedding feature $\mathbf{f_v}$ through multiple supervisions.
%The motion of dynamic objects can alter shadows and lighting, which often introduces artifacts in static regions during inpainting. 
To introduce a descriptor that better represents target region attributes under different dynamic shadows and lighting, we embed visual features $f$ into the neural Gaussians. Inspired by feature-embedded GS~\cite{huang2025stdloc, zhou2024feature}, we construct the visual feature embedding by jointly optimizing both radiance and feature fields through multi-task supervision. Fig.~\ref{fig:feature} shows the basic method.
We denote $ F_t(I) \in \mathbb{R}^{H' \times W' \times D} $ as the ground truth feature map obtained from the visual feature extractor (e.g., Superpoint~\cite{detone2018superpoint}), where $D$ is the visual feature dimension and $I \in \mathbb{R}^{H \times W \times 3}$ is the training image. Similar to the image rasterization method of Gaussian Splatting, we render the embedded visual feature $f$ from the perspective of the training camera and obtain the rendered feature map $\hat{F}_s(I)$. The visual feature loss $\mathcal{L}_{\mathrm{Feat}}$ is computed as the L1 distance between rendered and ground truth features:

\begin{equation}
\mathcal{L}_{\mathrm{Feat}} = M_i \cdot \|F_t(I) - \hat{F}_s(I)\|_1,
\label{eq:feat-loss}
\end{equation}

where $M_i$ is the dynamic object mask to eliminate the interference of dynamic objects on the visual feature extraction of static background neural Gaussians. %While derived from multi-view observations, \ssh{our visual feature} embeddings are intrinsically defined in 3D space to maintain consistency across viewpoints, with the anchors further enhancing local Gaussian aggregation for efficient patch retrieval during scene manipulation.

\textbf{Neural Gaussian Reconstruction.}
We reconstruct the driving scenes $\mathcal G$ via Neural Scene Graphs (NSG)~\cite{ost2021neural} with visual feature embedded anchors $\mathbf{p}\in\mathcal{G}$ hierarchically organized by Scaffold-GS\cite{lu2024scaffold}.  During the preparation phase for 3D Gaussian scenes before editing, we begin by initializing the 3D Gaussians using Structure-from-Motion (SfM) and LiDAR-stitched point clouds~\cite{yan2024street}. Subsequently, we refine the Gaussian attributes along with a visual feature embedding for each neural Gaussian through training, utilizing the following loss function:

\begin{equation}
\mathcal L=((1-\lambda_{\mathrm{SSIM}})\mathcal L_1 + \lambda_{\mathrm{SSIM}}\mathcal L_{\mathrm{SSIM}} + \lambda_{\mathrm{Depth}}\mathcal L_{\mathrm{Depth}}+\lambda_{\mathrm{Feat}}\mathcal{L}_{\mathrm{Feat}}),
\label{eq:total-loss}
\end{equation}

% \begin{equation}
%     \mathcal L_1 + \mathcal{L_{\mathrm{Feat}}}
% \end{equation}
%
where $\mathcal L_1$ and $\mathcal L_{\mathrm{SSIM}}$ are the L1 and SSIM losses between the rendered image and the ground-truth image, following the original 3DGS \cite{Kerbl20233dgs}. $\mathcal L_{\mathrm{Depth}}$ is the L1 loss between the rendered depth and the projected LiDAR depth. $\mathcal L_{\mathrm{Feat}}$ is the supervision for visual feature embedding as described in Eq.~\ref{eq:feat-loss}.
$\lambda_{\mathrm{SSIM}}$, $\lambda_{\mathrm{Depth}}$ and $\lambda_{\mathrm{Feat}}$ are hyper-parameters for balancing the terms. Note that the visual feature embedding is inferred from a different anchor feature than other Gaussian attributes, resulting in separate gradient backpropagation paths. %Consequently, training the visual feature embedding does not impact the reconstruction quality.

\subsection{Locating Target Patches}

\label{sec:method:locate}
Based on the visual feature embedded neural Gaussian representation, the first primary task is to locate the objective 3D regions to be inpainted. We define a `target patch' $\mathbf{T}_i$ as a continuous group of anchors, where some part within this patch should be removed, and the other parts can be regarded as hints for contextual searching.
% Furthermore, tactically, we build an efficient searching-and-indexing data structure to support the subsequent candidate patches searching and measuring.
% Here, we first describe how we locate the candidate target patch, and thus identify the object regions to be inpainted within the target region.
%need to accurately identify specific 3D region to enable further} 3D editing.
%in 3D editing tasks for driving scenarios, the main objectives involve the removal, replacement, or insertion of movable objects. Consequently, the primary task is to locate and select the regions to be edited. 

\textbf{Target Patch Identification.} We first propose to automatically locate target Gaussian regions mainly by utilizing the semantic masks of target objects to identify the corresponding anchors within the reconstructed scene. Specifically, we back-project the semantic mask to anchors during the neural Gaussian reconstruction and identify anchors with low opacity (alpha < 0.9) as inpainting targets. Low opacity represents unconverged Gaussians during training, which should be targeted. The selected anchors are then gathered as the target patch.
Optionally, users can manually select target patches for flexibility. 

% Note that if a frame contains multiple objects to be removed, we select and edit them in an order from nearest to farthest. 

% \chenming{Add some sentences to introduce the following contents, otherwise it is hard to read.}
\textbf{Classification of Anchor Points in Target Patch.} 
%To achieve a more precise comparison and replacement at the anchor level, we classify the anchors $\mathbf{p}$ within the target region based on whether the attributes of neural Gaussians in these regions are compensated sufficiently from affected view directions.
Thereafter, we propose to classify anchors $\mathbf{p}$ within the target patch into three categories as shown in Fig.~\ref{fig:pipeline} -- missing, incomplete, and intact -- for clearly separating the \emph{objectives} and the \emph{reliable searching references}.
% evaluating the \emph{completeness} of patches after reconstruction. %so as to add additional targets for maintaining quality. Regarding completeness,
% \ssh{Here since the} intact anchors \ssh{are often reconstructed with} reliable appearance and geometry, \ssh{we didn not perform inpainting for the intact anchors but using them to} find suitable candidates. Whereas, incomplete \ssh{and missing} anchors are unreliable parts to be \ssh{further} substituted. \sh{We should talk `Here since' before `Thereafter'.}
% \chenming{Why we need to define completeness here? shouldn't we have some sentences to introduce the requirements?}
%To obtain the target region and category of contained anchors, we first locate them by 2D-3D back-projection. Then, 
Specifically, we select the frame that contains the specified target object and extract \emph{missing} areas by picking opaque areas of the rendered alpha map from the viewpoint of the selected frame after removing the target objects. Then, we separate \emph{incomplete} from \emph{intact} anchors by projecting all anchor positions onto the preceding and succeeding frames of the previously selected frame and evaluate the rate of recurrence $\lambda_{\mathrm{CMP}}$ when they fall within the 2D semantic mask of the specified targets. 
These \emph{incomplete} anchors lacking supervision are imperfectly reconstructed parts, and thus, they are not further used for contextual candidate source patch searching.
% In this manner, we can obtain a set of anchors within the target area that lack supervision. 

% \textbf{Spatial Voxel-Based Scene Partitioning Structure.} 
\textbf{Volumetric Bidirectional Patch-Anchor Indexing.}
Once the target patches are identified with their anchors classified, we create bidirectional indexing between patch and its anchors for efficiency.
Specifically, given the spatial extent of a patch with a voxel size parameter $\lambda_\mathrm{size}$, we construct a corresponding voxel grid in the 3D space with hash-encoded bidirectional indexing between anchors and voxels.

% \sh{What is the relation between this proposed data structure and `Locating Target Patches'? We probably need a figure here, and thorough ablations on advantages of leveraging spatial partitioning.}

\subsection{Searching and Measuring Candidate Patches}
Our next goal is to find the candidate source patches $\mathbf{S}_j$ in 3D space to inpaint the target region. We further describe our searching and measuring strategy.

\textbf{Manifold BEV Searching.}
Unlike 2D patches searching within images, the 3D patches searching can be more ambiguous, which costs higher time complexity. For an efficient 3D patch searching within the driving scene reconstruction, our key observation is that the most suitable candidate patches are typically found near the curved surfaces where the target patch resides, particularly in road scenes where the surface can be modeled as a 2D manifold. Moreover, the longitudinal and lateral directions of lanes have the higher probability of obtaining best-matched source patches. This motivates us to introduce a manifold BEV patch searching.

Specifically, we first decompose the ground manifold into two main directions w.r.t. road elements and the driving direction of the raw clip, and then construct a BEV space orthogonal to the manifold surface as the basic searching space. 
Given a target patch $\mathbf{T}_i$ to be substituted, we perform a bidirectional path searching along the X-axis in the orthogonal manifold BEV space and then construct a rectangular area on the manifold for searching. 

\textbf{Patch Affinity.} 
During the manifold BEV searching for each target region, we also build the indexing structure to the neighborhood around the target region and then perform patch affinity measurement on the anchor level. 
Subsequently, for each voxel $\mathbf{V}^{\mathcal{T}}_m$ in the target patch $\mathbf{T}_i$, we iterate over the voxels $\mathbf{V}^{\mathcal{S}}_n$ in the source patches $\mathbf{S}_j$, extracting the anchors from both voxels for measurement. We only use \emph{intact} anchors for similarity comparison. Specifically, we separately project the center of each anchor in voxels $\mathbf{V}^{\mathcal{T}}_m$ and $\mathbf{V}^{\mathcal{S}}_n$ onto the image plane, generating two distinct 2D coordinate patches: $ \pi(\mathbf{T}_i) $ for $\mathbf{V}^{\mathcal{T}}_m \in\mathbf{T}_i $ and $ \pi(\mathbf{S}_j) $ for $ \mathbf{V}^{\mathcal{S}}_n\in\mathbf{S}_j $, where the corresponding pixel positions are set to 1. As Fig.~\ref{fig:feature} shows, we then extract the corresponding visual features from the rendered feature map $ \hat{F}_s(I)$ to compute the cosine similarity between patches as follows:
\begin{equation}
\mathbf{A}_{i, j} = \left\langle \hat{F}_s(I)[\pi(\mathbf{T}_i)], \hat{F}_s(I')[\pi(\mathbf{S}_j)] \right\rangle,
\label{eq:aff}
\end{equation}
where $\hat{F}_s[\pi(\mathbf{T}_i)]$ and $\hat{F}_s[\pi(\mathbf{S}_j)]$ represent the bilinearly interpolated visual features extracted from $\hat{F}_t$ at the projected regions $\pi(\mathbf{T}_i)$ and $\pi(\mathbf{S}_j)$, respectively. We specifically select the rendered feature maps from the viewpoints closest to the target region, denoted as $\hat{F}_s(I)$ for the target and $\hat{F}_s(I')$ for the source feature extraction. Note that $I'$ and $I$ may correspond to different frames or can be the same frame as well. The optimal source patch $\mathbf{S}^*_j$ is obtained by $\mathbf{S}^*_j \leftarrow \mathrm{argmax}_j(\mathbf{A}_{i,j})$, which will serve as the final selection.

\label{sec:method:affinity}

% \subsection{Placing and Blending Patches}
% \label{sec:method:blending}

% Once we have found such an appropriate candidate $\mathbf{S}_j$ for $\mathbf{T}_i$, we directly copy attributes for missing anchors, and blend attributes for partially reconstructed anchors. Specifically, we \shc{xxx}. \sh{Think how. And we also need to find out if there is anything to do with upper-level node(s).}
% %
% \begin{equation}
% \mathbf{B_{i,j}} = \tv{\mathbf{S}_j},
% \label{equ:blending}
% \end{equation}
% %
% where \shc{xxx}.

% \sh{Probable advantages of directly blending in 3D instead of 2D. Discuss after these two sub-sections proposed here: ensuring geometric consistency, and performing reliable operations on spherical harmonics for view-depend assets inpainting.}

%\subsection{Coarse-to-Fine Placing through Levels-of-Details}
\subsection{Substitution and Fusion}

\label{sec:method:fusion}

\begin{figure}[htb]
\centering
    \includegraphics[width=1.0\linewidth]{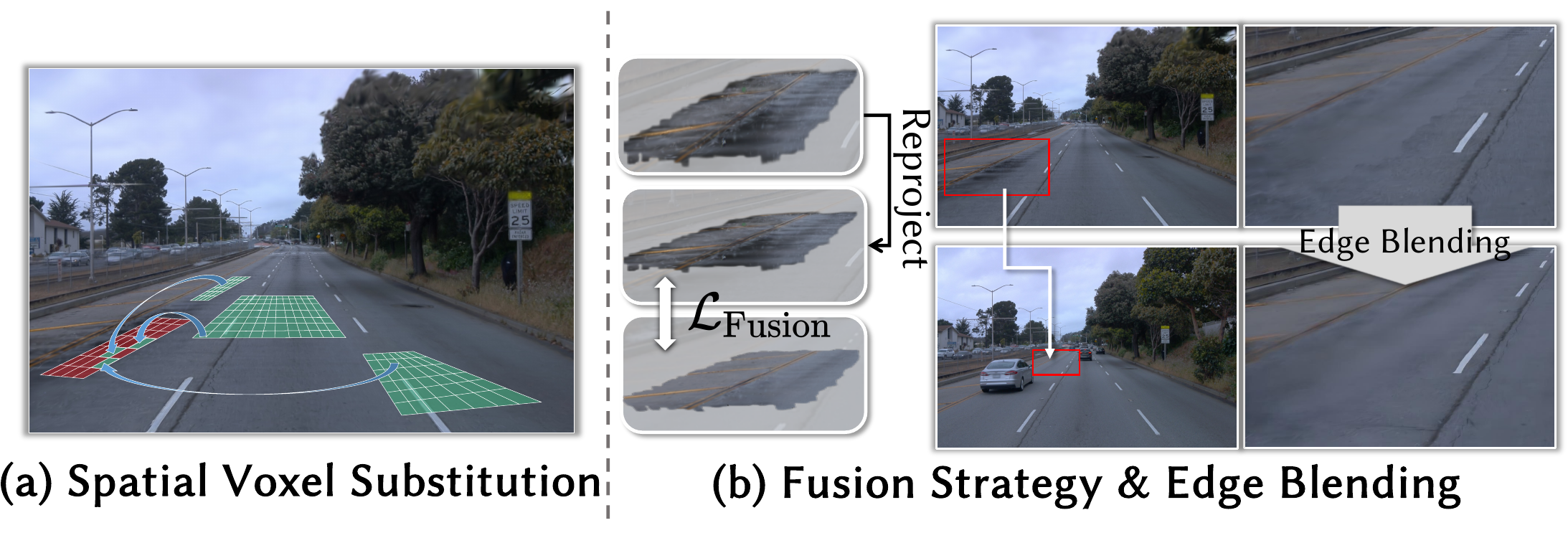}
    \caption{Illustration of substitution and fusion. We first perform (a) voxel-based patch substitution and then do (b) reprojection-guided fusion with edge blending for better visual harmony and multiview consistency.}
    \label{fig:fusion}
\end{figure}

After finding the optimal source patch $\mathbf{S}^*_j$ for each target patch $\mathbf{T}_i$, we conduct preliminary inpainting by transferring neural Gaussians in anchor-based units. To ensure proper blending, we set the opacity of neural Gaussians outside the target area to zero, based on their anchor positions and spatial offsets.
While the voxel-based anchor substitution can effectively reduce artifacts in the target region, visual inconsistencies still remain. To address this problem, we introduce a reprojection-guided fusion strategy, with a blending method applied specifically to the edge regions, achieving better visual harmony and multi-view and temporal consistency. The whole process is shown in Fig.~\ref{fig:fusion}.

Our key insight stems from the observation that dynamic foreground objects create temporal windows where background regions become visible before being occluded. Therefore, we can leverage this as a natural reference for spatial-temporal consistency. We extract the clean RGB background patch $\hat{B}$ from those unoccluded frames (i.e., reference viewpoints) and reproject it to the viewpoints where they become obstructed. The reprojected patch is then stitched with the rendered image. To reduce the visual discrepancy at the boundary between the reprojected patch and the rendered image $\hat{I}$, we apply dilation and erosion processes to the reprojected patch edges and perform alpha-blending on the edges with the rendered image to smooth out the visual discrepancy, which can be expressed as:
\begin{equation}
\hat{I}' = W_{\alpha} \cdot \pi(\hat{B}) + (1 - W_\alpha) \cdot \hat{I},
\end{equation}
where $\pi(\hat{B})$ refers to the reprojected background patch from the reference viewpoint to the target viewpoint, and $W_\alpha$ is the alpha-blending weighted mask. In practice, a fixed value $W_\alpha = 0.5$ is applied along edges with a bandwidth of 10 pixels. This approach allows us to generate some synthetic ground truth RGB images $\hat{I}'$ , then we use an L1 loss similar to the reconstruction process to fuse the target region, which can be expressed as:
\begin{equation}
\mathcal{L}_\text{Fusion} = \frac{1}{N} \sum^{N} \| \hat{I}' - \hat I \|_1.
\end{equation}
The fusion process relies exclusively on the RGB supervision, which ensures that the inpainted target region visually aligns with its neighboring areas. As a result, effective optimization in terms of temporal and multi-view consistency can be achieved through supervised updates with very few iterations.

In summary, Substitution initializes with geometry-aware patch transfer, Fusion enforces global visual consistency, and Edge-blending ensures smooth transitions along patch boundaries. This design ensures high-quality and coherent scene completion across views and time.
% clarification on three components

\section{Implementation, Result and Evaluation}
\label{sec:exp}

% \begin{figure*}[t]
%     \centering
%     \includegraphics[width=\linewidth]{figures/full.pdf}
%     \caption{Qualitative comparison with baseline inpainting methods on the Waymo Open Dataset~\cite{waymodataset}. Specified targets (in red boxes) are removed through baseline and ours methods. Zoomed-in figures illustrate our effectiveness of harmonious removal.}
%     \label{fig:full_comparsion}
% \end{figure*}

\subsection{Experimental Setup}
\label{sec:exp:setup}

\textbf{Implementation Details.}
We train each sequence for 30,000 iterations using the Adam optimizer, with the hyperparameters set to $\lambda_{\mathrm{SSIM}} = 0.2$ and $\lambda_{\mathrm{Depth}} = 0.2$, consistent with the Scaffold-GS~\cite{lu2024scaffold} and $\lambda_{\mathrm{Feat}}=1.0$, refer to STDLoc~\cite{huang2025stdloc} . We choose  $\lambda_{\mathrm{CMP}} = 0.5$, and $\lambda_{\mathrm{size}}=250cm$ and refer readers to Sec.~\ref{sec:exp:ablation} for detailed ablation studies. 
All experiments for both our proposed method and the baseline methods were performed on an NVIDIA RTX 3090 GPU.

\textbf{Datasets.}
To evaluate the effectiveness of the proposed method for inpainting static street scenes in driving scenarios, we conduct experiments using the Waymo Open Dataset~\cite{waymodataset}.
Specifically, we select 23 sequences, each containing approximately 60$\sim$90 frames, including 7 publicly tagged multi-dynamic object scenes, 4 severely occluded scenes, 6 uphill and downhill scenes, as well as 4 scenes under complex weather and lighting conditions. We use the object tracking results originally provided by the dataset itself to specify target regions. Tab.~\ref{tab:dataset} shows detailed sequence information.

\begin{table}[htbp]
    \centering
    \caption{Detailed selected sequences from the Waymo Open Dataset~\cite{waymodataset}.}
    \fontsize{8pt}{9.6pt}\selectfont {\begin{tabular}{l|c}
    \toprule
       Type  &  Sequence ID. \\
    \midrule
       plain road  & 031, 084, 089, 592, 640 ,796   \\
       multi-dynamic object & 023, 024, 034, 058, 108, 115, 438  \\
       up-down hill & 124, 147, 206, 382, 546, 574   \\
       complex weather/lighting & 044, 124, 147, 574\\
       severe occlusion & 001, 021, 115, 592\\
    \bottomrule
    \end{tabular}}
    
    \label{tab:dataset}
\end{table}

\textbf{Baseline Selection.} For driving scenes, we compare our methods with three publicly available approaches. Two of them are two-stage GS-based inpainting methods: GScream~\cite{wang2025gscream}, InFusion~\cite{liu2024infusion}. Apart from that, we also compare our method with one of the advanced 2D video inpainting methods, ProPainter~\cite{zhou2023propainter}. In addition, due to the fact that the reconstruction methods of baseline schemes often degrade in street scenes, we also applied the InFusion method to the background Gaussians based on the baseline StreetGaussians~\cite{yan2024street} of SoTA street reconstruction for fair comparisons.
We exclude VideoPainter~\cite{bian2025videopainter} due to its very high inference costs ($>40$GB VRAM) and limited relevance to our 3D-based editing framework. Additionally, StreetUnveiler~\cite{xu2024streetunveiler} necessitates diffusion-based inpainting for videos and relies heavily on extensive semantic segmentation data for reconstruction, leading to significant differences in input requirements. This makes fair comparisons impractical, so we also decide to omit it from our experiments. For details of the baseline experimental settings, please refer to the supplementary material.

\subsection{Result and Discussion}
\label{sec:exp:results}

We conduct qualitative and quantitative comparisons in terms of inpainting performance against three baseline methods. For detailed metric calculations and additional inpainting demonstrations, we refer readers to the \emph{supplementary materials}.

\begin{figure*}[htb]
    \centering
    \includegraphics[width=0.93\linewidth]{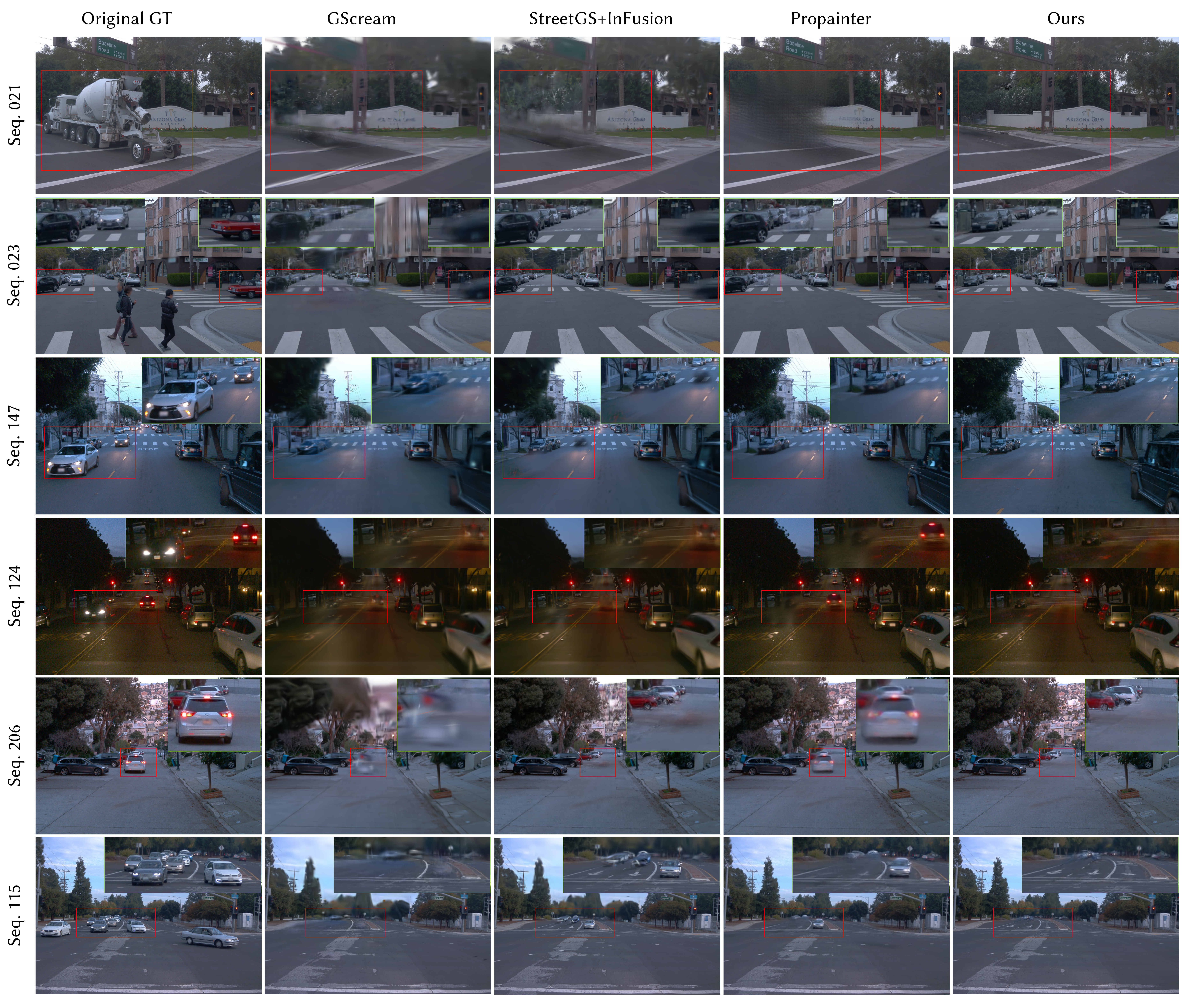}
    \caption{Qualitative comparison with baseline inpainting methods on the Waymo Open Dataset~\cite{waymodataset}. Specified targets (in red boxes) are removed through baseline and ours methods. Zoomed-in figures illustrate our effectiveness of harmonious removal.}
    \label{fig:full_comparsion}
\end{figure*}

\textbf{Inpainting Performances.}
As shown in Tab.~\ref{tab:main_exp}, we evaluate the LPIPS and FID scores on the rendered images, demonstrating that our method consistently outperforms all baseline GS-based inpainting methods across all metrics and tested sequences.
Fig.~\ref{fig:full_comparsion} presents a qualitative comparison across different driving scenes. For individual moving objects as shown in seq$124$ and seq$147$, our method demonstrates high-quality inpainting under various lighting conditions. In complex scenes with multiple moving objects, such as seq$023$ and seq$115$, our method successfully avoids artifacts and produces highly realistic removal results, while GScream and InFusion struggle to achieve clean foreground removal. Notably, despite ProPainter’s low LPIPS score since it directly performs on the original video sequence, it still inevitably produces significant artifacts as shown in Fig.~\ref{fig:full_comparsion}. Compared with video inpainting methods, our method enables geometry-aware 3D scene completion and editing across novel views, while remaining effective and surpassing baseline performance.
% Our method remains effective, surpassing baseline performance.
% \sh{Please check this core discussion para carefully, and check whether some discussions in the beginning sections (Intro + Rel) has been re-called here. @CG, @JR.}

In terms of processing efficiency, we also evaluate the inpainting time of different algorithms. InFusion demands an additional 150 training iterations along with 2D inpainting processing.
Our method only needs an additional 50 training iterations with post-fusion with no 2D inpainting prior. In contrast, GScream, which is more efficient than the other methods and ablations, includes only an additional 2D inpainting step and performs 3D inpainting during training. However, it produces relatively poor results. Other methods involve time-costly 2D prior generation, which heavily degrades the efficiency of inpainting.

\begin{table}[htbp]
    \centering
    \caption{Quantitative results on Waymo Open Dataset~\cite{waymodataset}. The \textbf{best} and \underline{second-best} metrics are highlighted in \textbf{bold} and \underline{underline}, respectively. We report \textbf{Time$\downarrow$} for the average extra processing time measured in seconds.}
\fontsize{8pt}{9.6pt}\selectfont {    \begin{tabular}{l|ccc}
    \toprule
        Method &  \textbf{LPIPS$\downarrow$} & \textbf{FID$\downarrow$} & \textbf{Time$\downarrow$} \\
    \midrule
    ProPainter~\shortcite{zhou2023propainter} & \textbf{0.207} & \underline{81.925} & 147 \\
    GScream~\shortcite{wang2025gscream} & 0.421 & 152.879 &  \textbf{20} \\
    InFusion~\shortcite{liu2024infusion} & 0.396 & 126.556 &  96 \\
    StreetGS~\shortcite{yan2024street}+InFusion & 0.243 & 114.490 & 115 \\
    \midrule
    Ours & \underline{0.237} & \textbf{74.124} & \underline{64} \\
    \bottomrule
    \end{tabular}}
    \label{tab:main_exp}
\end{table}

\begin{figure}[htb]
\centering
    \includegraphics[width=0.98\linewidth]{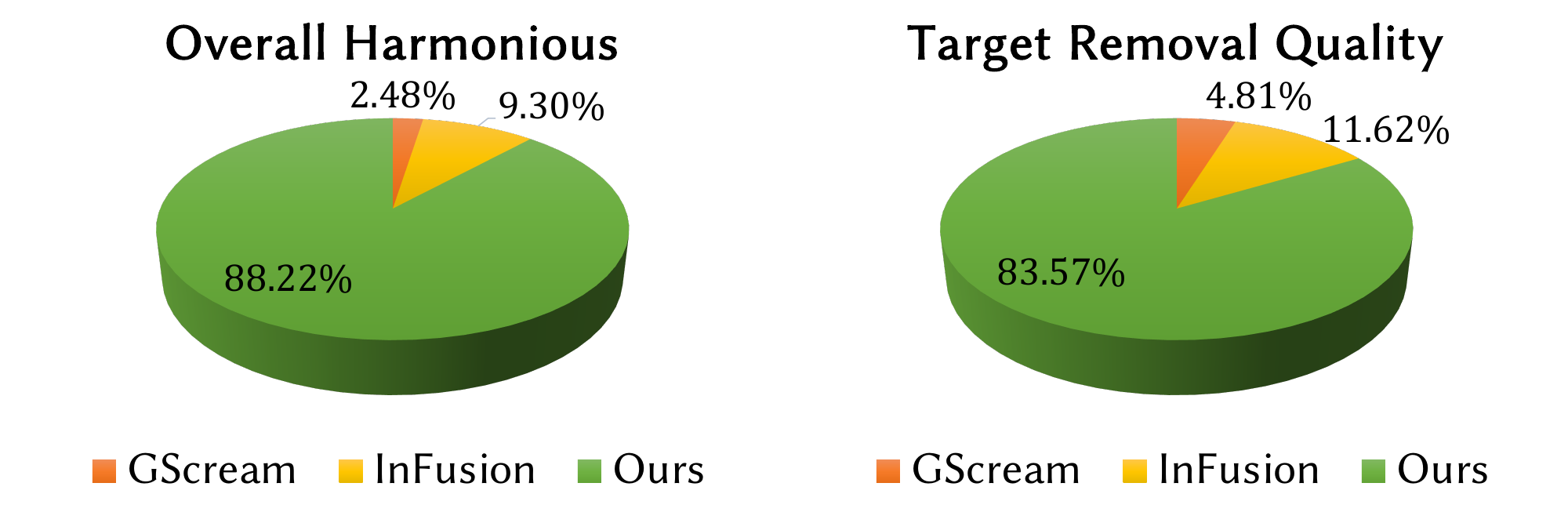}
    \caption{User study of object removal. We recorded the proportion of participants choosing for each scenario.}
    \label{fig:user}
\end{figure}

\textbf{User Study.} We conduct a user study to assess the harmony of our scene editing results using the outputs shown in Fig.~\ref{fig:full_comparsion}.  We recruited 43 data labelers with varying levels of experience in labeling images and videos for autonomous driving tasks. To evaluate the results, we prepared 3 groups of video sequences and 12 groups of image frames, with each group containing three randomly shuffled outputs from GScream, InFusion, and our method. Participants are independently asked to select the best option among the three choices based on the following two criteria: (1) ``Given three choices, which one has the most natural quality as real-world shots?'' and (2) ``Given three choices and an additional original real-world shot as a reference, which one has achieved a clean removal of the specified target?''
Fig.~\ref{fig:user} summarizes the distribution of choices from these participants, which again indicates that our method has achieved a better visual harmony when compared to other baseline methods.
As a result, 88.22\% and 83.57\% of participants across all cases select our inpainted results as the best, further demonstrating that our method achieves superior qualitative performance compared to other baseline methods. This user study was approved by the Institutional Review Board (IRB).

\textbf{Expansion on Static Scene Inpainting.} To further demonstrate the universal application value of our proposed inpainting strategy in 3D space, we conduct additional comparative experiments on 360-USID dataset~\cite{wu2025aurafusion360}. This dataset includes seven $360^\circ$ unbounded scenes with training views and validation views without objects for testing. The dataset was designed for reference-based inpainting methods, yet we conduct our method without any ground truth reference views. For other methods ~\cite{mirzaei2023spin, ye2025grouping, wu2025aurafusion360}, we follow the baseline results of the officially released code base. Tab.~\ref{tab:usid} reports the evaluated PSNR, SSIM and LPIPS metrics compared against the validation views. The results demonstrate that our method not only generalizes well to diverse non-street scenarios but also outperforms existing NeRF-based and 2D prior-based inpainting approaches.

\begin{figure*}[hb]
    \centering
    \includegraphics[width=1.0\linewidth]{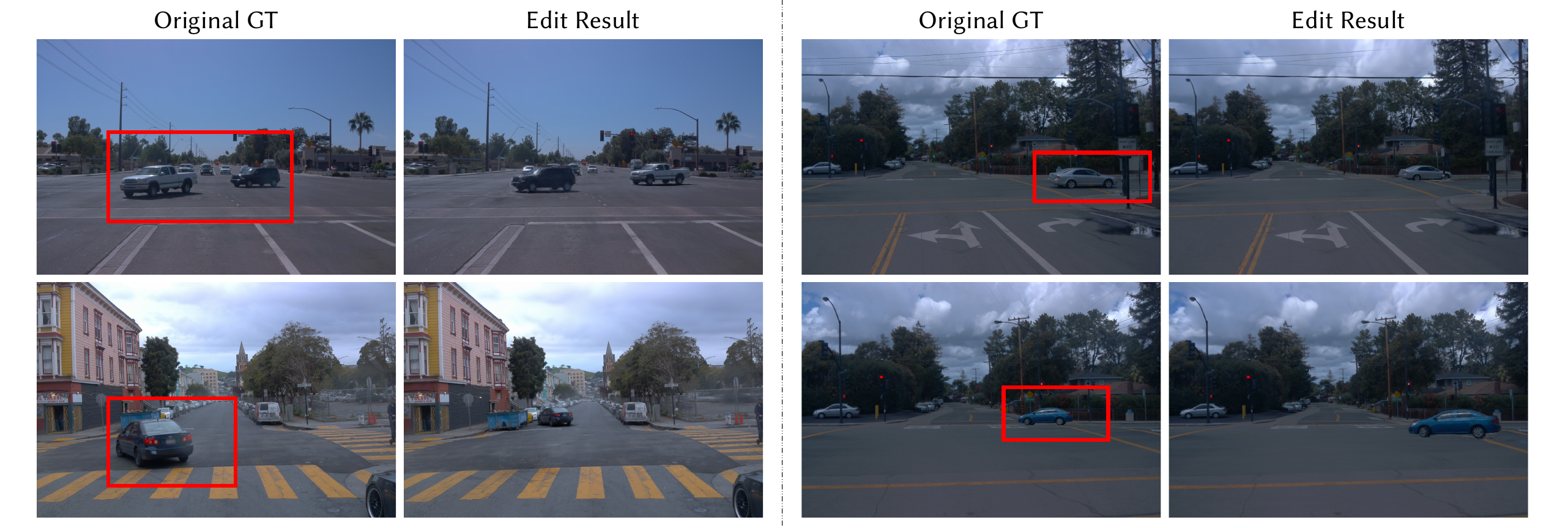}
    \caption{Scene editing examples utilizing our inpainted driving scenes. Our proposed GS-RoadPatching directly inserts or moves the foreground vehicles on reconstructed 3D assets for harmonious rendering and driving clip variation.}
    \label{fig:edit}
\end{figure*}

\textbf{Applications.}
The proposed method is applicable to a range of driving scene tasks, including reconstruction, object removal, and scene-level editing. As illustrated in Fig.~\ref{fig:teaser}, Fig.~\ref{fig:edit}, and the supplementary video, our approach facilitates visually coherent reconstruction and 3D-aware editing of driving scenes.

\begin{table}[htbp]
    \centering
    \caption{Quantitative results on 360-USID dataset, all metrics are computed for the target region. The \textbf{best} and \underline{second-best} metrics are highlighted in \textbf{bold} and \underline{underline}, respectively.}
    \fontsize{8pt}{9.6pt}\selectfont {\begin{tabular}{l|ccc}
    \toprule
         Method &  \textbf{PSNR$\uparrow$} & \textbf{SSIM$\uparrow$} & \textbf{LPIPS$\downarrow$} \\  
    \midrule
       SPIn-NeRF~\shortcite{mirzaei2023spin}  & 16.734 & 0.958 & 0.464 \\
       GaussianGrouping~\shortcite{ye2025grouping} & 16.074 & 0.954 & 0.480 \\
       GScream~\shortcite{wang2025gscream}  & 14.758 & 0.956 & 0.514  \\
       InFusion~\shortcite{liu2024infusion}  & 14.416 & 0.955 & 0.484  \\
       AuraFusion360~\shortcite{wu2025aurafusion360}  & \underline{17.661} & \underline{0.960} & \underline{0.388}\\
       \midrule
       Ours & \textbf{17.917} & \textbf{0.975} & \textbf{0.360} \\
    \bottomrule
    \end{tabular}}
    \label{tab:usid}
\end{table}

\subsection{Ablation Study}
\label{sec:exp:ablation}
This section provides ablation studies on our method, and the \emph{supplementary material} contains multiple sequences to demonstrate the effectiveness of terms and strategies.

% \textbf{Splitting intact and incomplete anchors.} For $\lambda_{\mathrm{CMP}}$ in Sec.~\ref{sec:method:locate}, we conduct ablation studies to determine a general value suitable for driving scenarios. In the experiments, we rank all points based on their proportion falling within the 2D semantic mask of specified targets in the frames preceding and succeeding the selected frame for projection. The points with the top $\lambda_{\mathrm{CMP}}$ values are identified as incomplete anchors lacking supervision.

% As illustrated in Fig.~\ref{fig:param_study}, a $\lambda_{\mathrm{CMP}}$ value of approximately $0.5$ delivers relatively satisfactory inpainting performance across most driving scenes. However, it is important to emphasize that this value may not be universally optimal. In extreme conditions--such as nearly static scenes with minimal vehicle movement or high-speed scenarios with limited supervision--further parameter adjustments may be necessary.

\textbf{Analysis of Scene Organization.} We conduct ablation studies on different voxel sizes during spatial indexing and patch selections. Our analysis reveals a trade-off: smaller voxels would lose the ability to maintain overall target region consistency during searching, whereas larger voxels amplify noise artifacts during replacement. The results indicate that patch selection and replacement based on a voxel size of 250 cm achieve optimal inpainting performance in most street scenarios.

\begin{table}[htbp]
    \centering
    \caption{Ablation studies on scene voxel size. The \textbf{best} and \underline{second-best} metrics are highlighted in \textbf{bold} and \underline{underline}, respectively.}
    \fontsize{8pt}{9.6pt}\selectfont {\begin{tabular}{c|cc}
    \toprule
       $\lambda_{\mathrm{size}}$(cm)  &  \textbf{LPIPS$\downarrow$} & \textbf{FID$\downarrow$} \\
    \midrule
       50  & 0.329 & 156.422 \\
       150 & 0.253 & 92.841  \\
       250 & \textbf{0.237} & \textbf{74.124}  \\
       350 & \underline{0.243} & \underline{84.341}  \\
       500 & 0.402 & 172.842  \\
    \bottomrule
    \end{tabular}}
    
    \label{tab:ablation_voxel}
\end{table}

% \begin{figure}[htb]
%     \centering
%     \includegraphics[width=0.6\linewidth]{figures/param_study.pdf}
%     \caption{Ablations on $\lambda_{\mathrm{CMP}}$ in Sec.~\ref{sec:method:locate}.}
%     \label{fig:param_study}
% \end{figure}

\begin{figure*}[htb]
    \centering
    \includegraphics[width=0.98\linewidth]{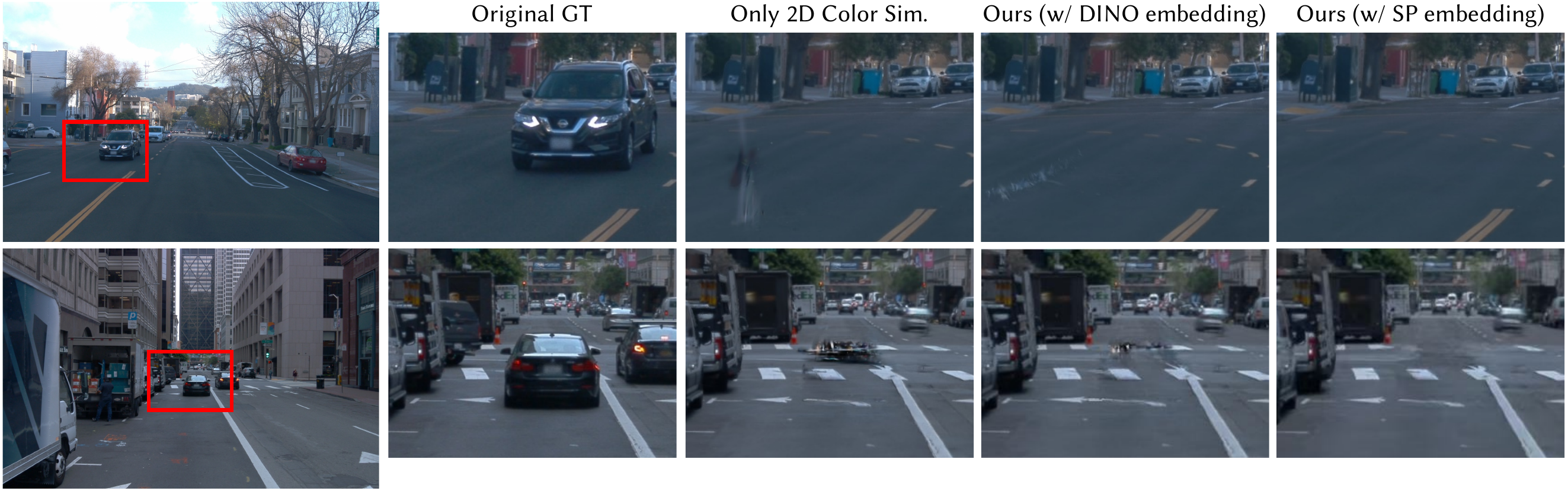}
    \caption{Qualitative results of ablation studies on feature embedding. We abbreviate SuperPoint as SP. DINO features are less robust and viewpoint-invariant than SP, which may find incorrect candidate patches.}
    \label{fig:ablation_measure}
\end{figure*}

\textbf{Analysis of Feature Embedding.} Tab.~\ref{tab:ablation_feature} compares the performance of different feature embeddings, including a baseline using simple 2D color similarity from projected patches. Our experiments show that SuperPoint’s lightweight features achieve superior cross-viewpoint consistency and robustness in matching. In contrast, DINO features~\cite{oquab2023dinov2} are more resolution-sensitive, exhibit weaker viewpoint invariance, and incur higher computational costs—both in training the feature embedding and computing patch affinities. The 2D color-only baseline performs poorly, as the target patch $\mathbf{T}_i$ often lacks discriminative color features, leading to noise artifacts during substitution that are unrelated to the target region. However, visual embeddings may degrade in sparsely supervised regions due to insufficient embeddings and texture learning.
Fig.~\ref{fig:ablation_measure} shows more visualized results.

\begin{table}[htbp]
    \centering
    \caption{Ablation studies on feature embedding. The \textbf{best} and \underline{second-best} metrics are highlighted in \textbf{bold} and \underline{underline}, respectively.}
    \fontsize{8pt}{9.6pt}\selectfont {\begin{tabular}{l|cc}
    \toprule
       Method  &  \textbf{LPIPS$\downarrow$} & \textbf{FID$\downarrow$} \\
    \midrule
       Ours (only 2D color sim)  & 0.354 &  92.453 \\
       Ours (DINO feature embedding) & \underline{0.258} & \underline{81.924} \\
    \midrule
       Ours (SuperPoint feature embedding) & \textbf{0.237} & \textbf{74.124} \\
    \bottomrule
    \end{tabular}}
    
    \label{tab:ablation_feature}
\end{table}

\begin{figure*}[htb]
    \centering
    \includegraphics[width=0.98\linewidth]{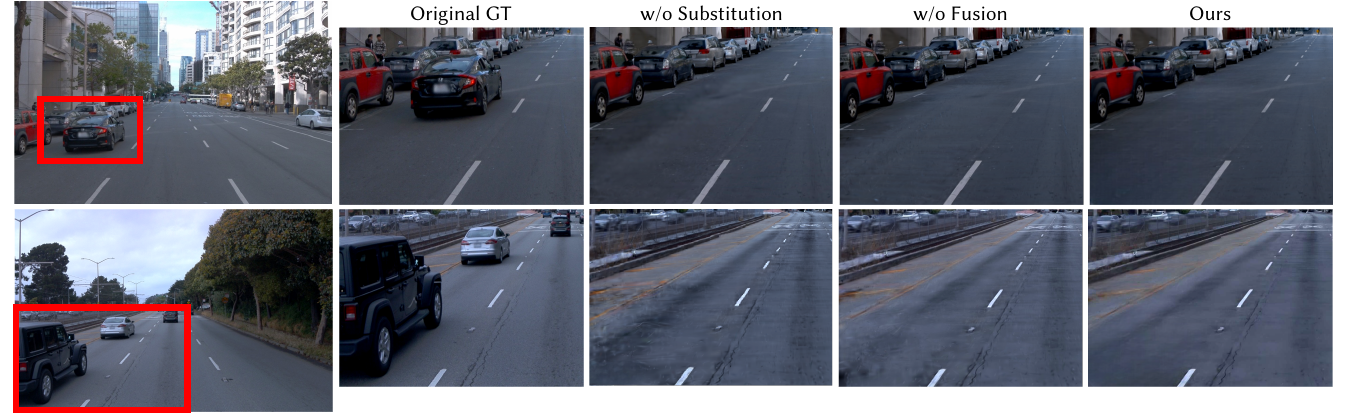}
    \caption{Qualitative results of ablation studies on fusion strategy in Sec.~\ref{sec:method:fusion}, reflecting our effectiveness of our proposed method.}
    \label{fig:ablation_fusion}
\end{figure*}

\textbf{Analysis of Substitution and Fusion Strategy.} Tab.~\ref{tab:ablation_fusion} shows results of inpainting on different fusion configurations. The results show that fusion enhances visual harmony more effectively than direct substitution. Ablating substitution leads to performance degradation, while edge blending under reprojection supervision provides additional improvement. Fig.~\ref{fig:ablation_fusion} shows more visualized results.

\begin{table}[htbp]
    \centering
    \caption{Ablation studies on fusion strategy. The \textbf{best} and \underline{second-best} metrics are highlighted in \textbf{bold} and \underline{underline}, respectively.}
    \fontsize{8pt}{9.6pt}\selectfont {\begin{tabular}{l|cc}
    \toprule
       Method  &  \textbf{LPIPS$\downarrow$} & \textbf{FID$\downarrow$} \\
    \midrule
       Ours (fusion only w/o substitution)  & 0.322 &  98.537 \\
       Ours (substitution only w/o fusion)  & 0.347 &  93.386 \\
       Ours (w/ fusion, w/o edge-blending)  &  \underline{0.263} & \underline{82.645}  \\
    \midrule
       Ours (substitution and fusion) & \textbf{0.237} & \textbf{74.124} \\
    \bottomrule
    \end{tabular}}
    
    \label{tab:ablation_fusion}
\end{table}

\section{Conclusion}
\label{sec:conclusion}

In this paper, we introduce GS-RoadPatching, an innovative inpainting method to remove dynamic objects in driving scenes, which is purely based on the search and editing of 3DGS scenes, without relying on any pre-edited 2D priors or guidance. Our method organizes the regions to be edited in the form of visual feature-embedded neural Gaussian patches within a patch-anchor indexing structure, and searches for candidate patches in the orthogonal manifold BEV space. We design a comparison strategy based on visual feature embeddings of neural Gaussian patches. Additionally, a substitution and fusion strategy is implemented, resulting in visually harmonious and temporally consistent inpainting results. 
We demonstrated the effectiveness of our method through experiments on diverse scenes in the Waymo datasets and further showed its applicability in more general cases. 
Future works may explore enhanced searching strategies and similarity measurements for object-level inpainting, or develop more efficient data structures for better locating and searching schemes in complex scenes. We hope that our work can inspire subsequent efforts for better inpainting based on the 3DGS reconstruction of driving scenes.

% \vfill

\textbf{Limitation.} % time and geometry
% Our method has been validated to achieve good inpainting results in most scenarios. 
While our method is suitable for high-quality offline scene completion or editing tasks, it is not designed for real-time deployment. Achieving real-time performance would require simplifying the algorithm or replacing the search and fusion process with lightweight predictive models. In scenarios where perspective supervision is severely limited (e.g., a large vehicle persistently blocks the view throughout the clip) or where the scene consists of distinct structures lacking reliable references, the patch searching and substitution process may be adversely affected.

\begin{acks}

We thank reviewers for their constructive comments and suggestions. This research was supported by the Zhejiang Provincial Natural Science Foundation of China under Grant No. LD24F030001. Shi-Sheng Huang was also supported by the National Key Research and Development Program of China under Grant No. 2024YFB2808804, and the Open Project Program of the State Key Laboratory of CAD\&CG (Grant No. A2402), Zhejiang University.

\end{acks}

% Bibliography
\bibliographystyle{ACM-Reference-Format}
\bibliography{main}

\begin{figure*}[htb]
    \centering
    \includegraphics[width=0.93\linewidth]{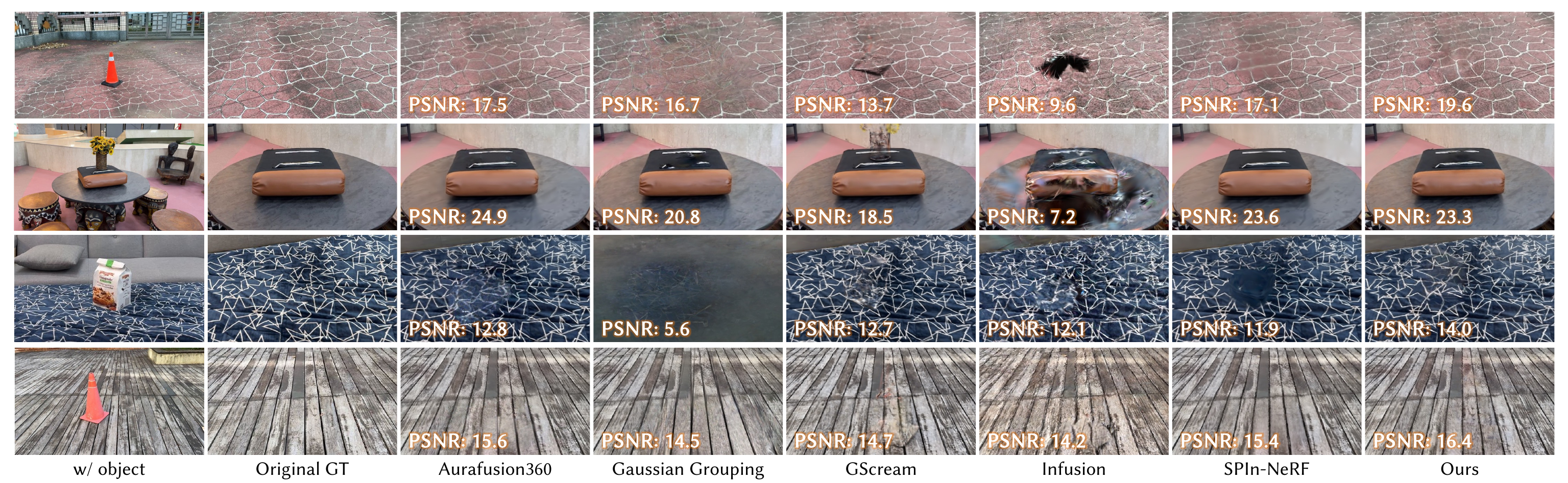}
    \caption{Qualitative results of ablation studies on 360-USID dataset~\cite{wu2025aurafusion360}.}
    \label{fig:full_usid}
\end{figure*}

\clearpage\clearpage
\appendix
\section{Experimental Details}
\textbf{Training Setups.} We follow the parameter setup as Scaffold-GS~\cite{lu2024scaffold}. For each attribute of neural Gaussians (i.e., color, rotation, scale and opacity), we use an corresponding MLP to learn its representation. As for the visual feature embeddings, we directly embed the 256-dimension SuperPoint~\cite{detone2018superpoint} feature into each neural Gaussians.
Specifically, we use Adam optimizer to train the scenes for 30,000 iterations, with learning rate of anchor feature, color, rotation, scale and opacity as $\eta_{\mathrm{ANF}}=0.0075, \eta_{\mathrm{CLR}}=0.008, \eta_{\mathrm{ROT}}=0.002, \eta_{{\mathrm{SCALE}}}=0.007, \eta_{\mathrm{OPA}}=0.02$. The neural Gaussian offset of each anchor is learned with an learning rate as $\eta_{\mathrm{OFFSET}}=0.01$. All learning rates are scheduled with a decay factor of 0.01. The scene is densified for 15,000 iterations, after that, we only update the existing anchors and prune the anchors with opacity less than 0.05.

% \textbf{Explanations on Dataset.}
% As mentioned in main text, we select sequences with multiple scene type from Waymo dataset. Specifically, \cg{058 for high-speed scenes, 044, 284 for complex lighting and weather scenes. }

\textbf{Baseline Implementations.} 
On the Waymo Open Dataset~\cite{waymodataset}, for GScream~\cite{wang2025gscream}, we use the bounding box of dynamic objects as a mask for training, and select a frame that contains a relatively complete dynamic object to be removed for 2D-inpainting. Following the experimental setup of GScream, we trained for 30,000 iterations and added cross-attention starting from the 15,000th iteration. 
For InFusion~\cite{liu2024infusion}, due to the poor performance of its vanilla 3DGS pipeline in autonomous driving scenes, we adopted the GScream pipeline without introducing cross-attention as the original scene training method for the first 30,000 iterations. Afterwards, we select and perform 2D color and depth inpainting proposed by InFusion on the selected output image, followed by 150 iterations of optimization. 
We adopts SDXL~\cite{podell2023sdxlimprovinglatentdiffusion} for generating 2D-inpainted result. The depth of the image and the depth of the inpainted image are obtained using Marigold~\cite{ke2023repurposing}.
On the 360-USID dataset~\cite{wu2025aurafusion360}, we follow the baseline settings introduced in AuraFusion360~\cite{wu2025aurafusion360} (see SuppDoc-F of AuraFusion360), where Gaussian Grouping~\cite{ye2025grouping} and our method do not rely on the reference image provided by the dataset, whereas other baselines leverage this reference image as an inpainting prior.

\textbf{Metrics.} For Waymo dataset, we calculate LPIPS and FID metrics by comparing the rendered images after inpainting with the original ground truth (GT) images. This approach has been used in some works to directly evaluate the effectiveness of inpainting~\cite{xu2024streetunveiler}. For 360-USID dataset, we calculate PSNR, SSIM and LPIPS compared to the ground-truth validation views.

\section{Additional Experiments}
\begin{figure*}[htb]
    \centering
    \includegraphics[width=0.95\linewidth]{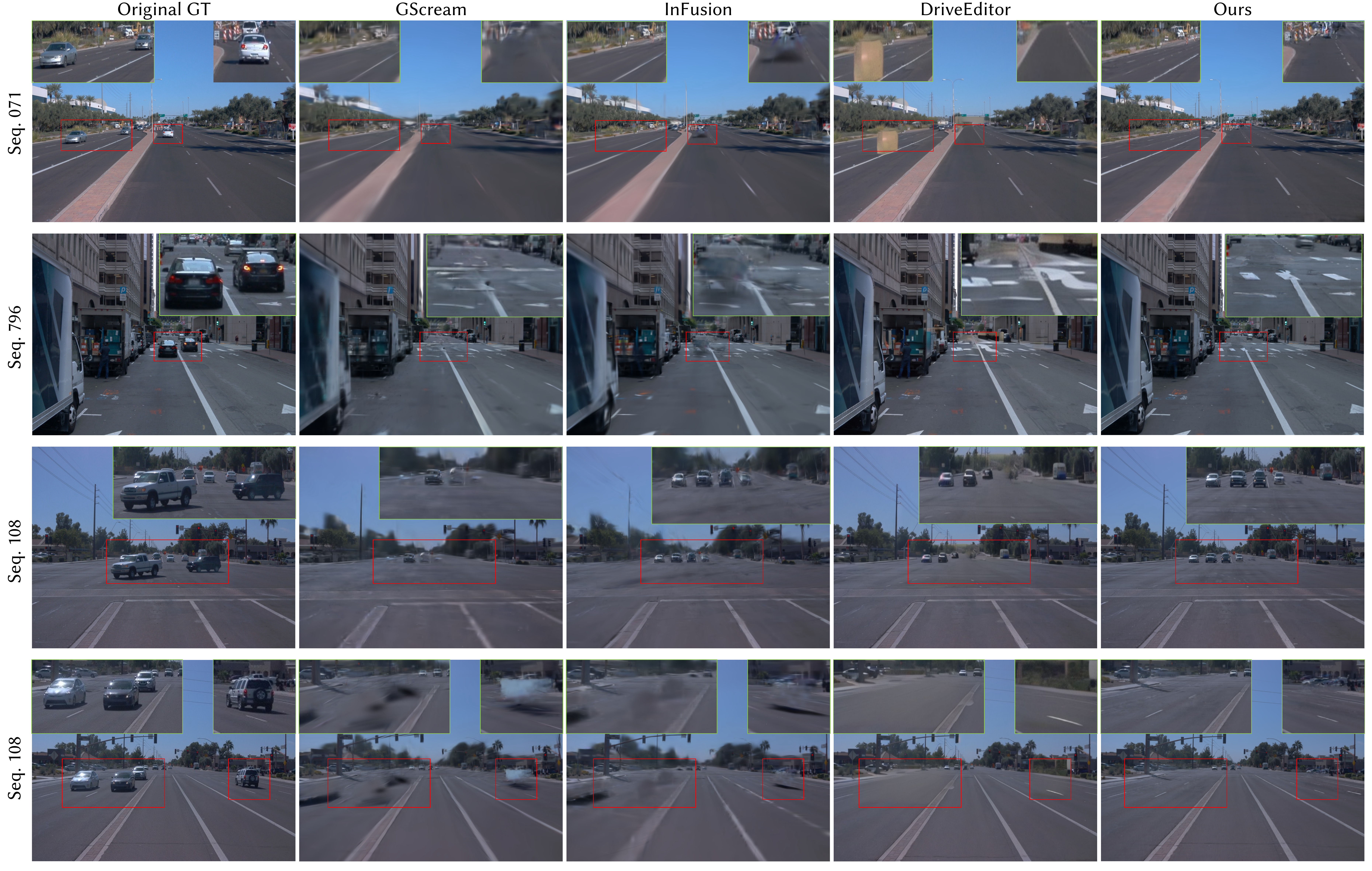}
    \caption{More qualitative comparison with baseline inpainting methods on the Waymo Open Dataset~\cite{waymodataset}. Specified targets (in red boxes) are removed through baseline and ours methods. Zoomed-in figures illustrate our effectiveness of harmonious removal.}
    \label{fig:full_additional}
\end{figure*}

Fig.~\ref{fig:full_additional} shows additional experiment results on more sequences in Waymo dataset~\cite{waymodataset}. Tab.~\ref{tab:supp_exp} further provides a quantitative comparison with video inpainting methods~\cite{zhou2023propainter, liang2025driveeditor}. The results demonstrate the effectiveness of our proposed method.

\section{User Study Details}
\begin{figure*}[htbp]
    \centering
    \includegraphics[width=\linewidth]{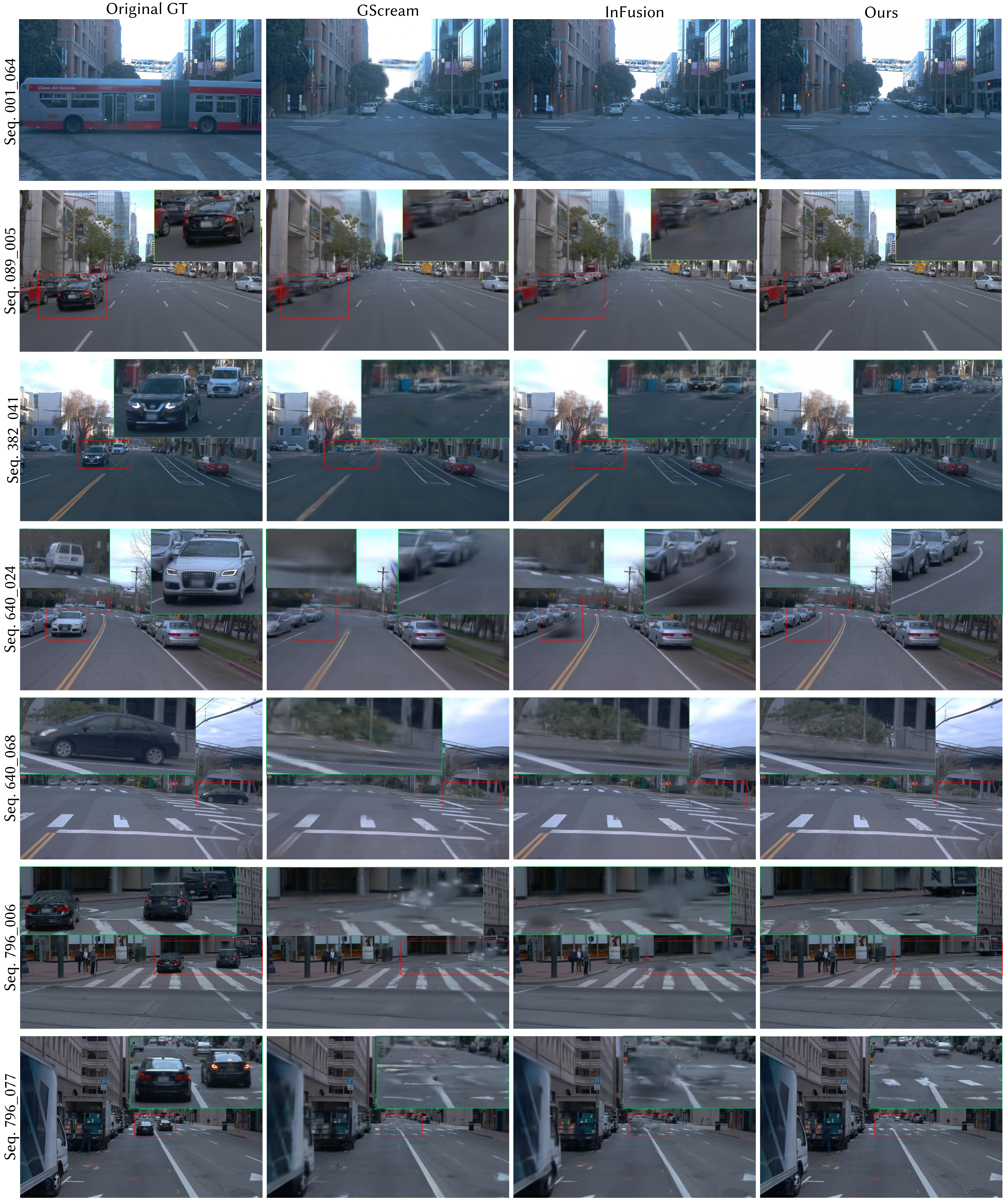}
    \caption{Qualitive results of images selected for user studies.}
    \label{fig:user_fig}
\end{figure*}

\begin{figure*}[htbp]
    \centering
    \includegraphics[width=\linewidth]{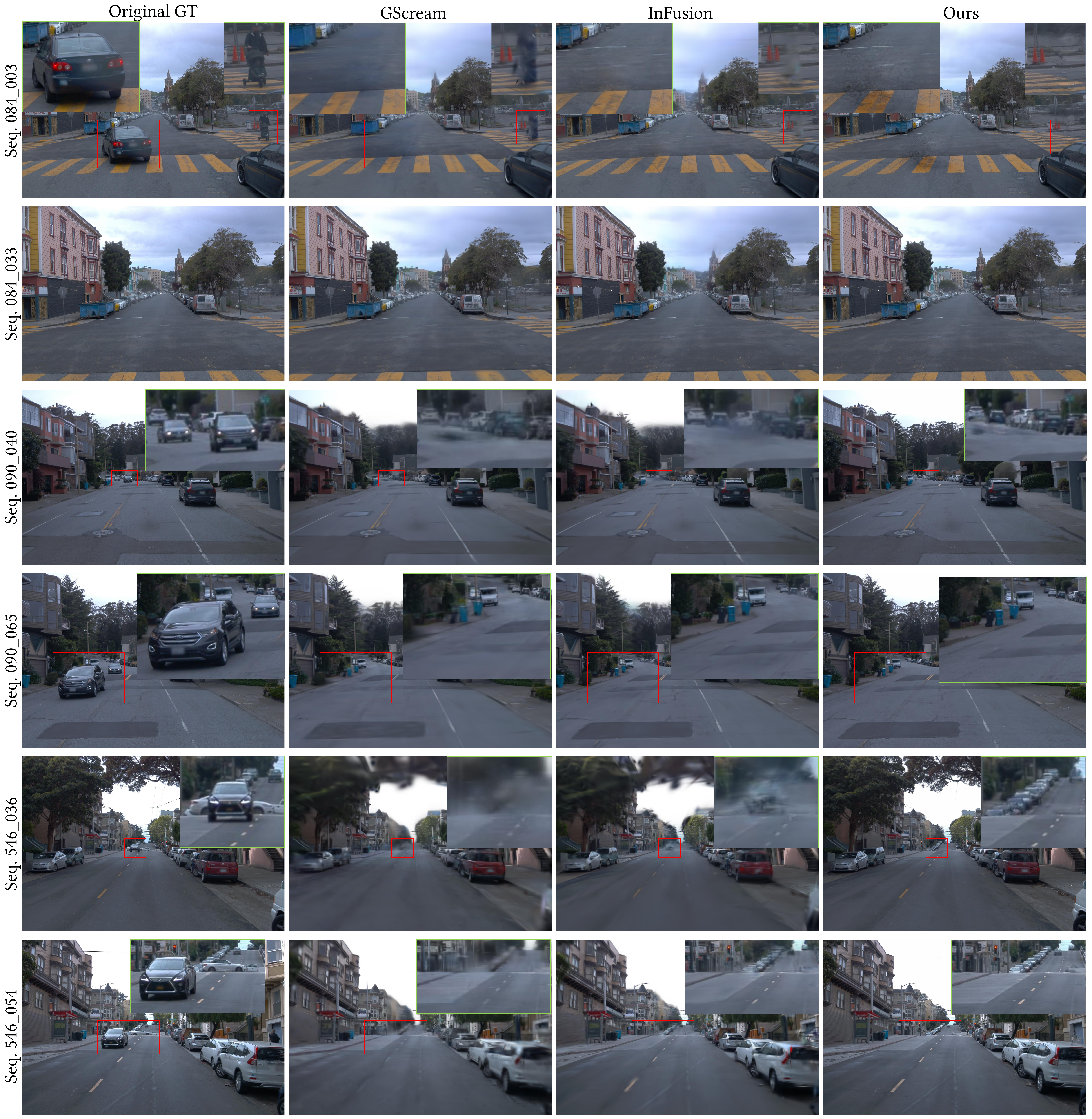}
    \caption{Qualitive results of video sequences selected for user studies. We show the start frame and end frame of each sequence.}
    \label{fig:user_vid}
\end{figure*}

Our user study included 43 participants, each responding to two questions regarding 12 groups of image comparisons and 3 groups of video sequence comparisons. Tab.~\ref{tab:user_full} presents the number of samples chosen as the best effect for each scheme. Fig.~\ref{fig:user_fig} and Fig.~\ref{fig:user_vid} illustrate some of the results outcome visuals for each option across the various schemes.

\begin{table*}[htb]
\caption{Statistics of voting for various choices in user study. We abbreviate Overall Harmonious as \textbf{OH} and Target Removal Quality as \textbf{RQ}, respectively.}
\centering
\resizebox{17.5cm}{!}{
\setlength{\tabcolsep}{3pt}
\begin{tabular}{l|cccccccccccccccccccccccc|cccccc|cc}
\toprule
\toprule
  \textbf{Seq\_Frm} & \multicolumn{2}{c}{\textbf{001\_064}} & \multicolumn{2}{c}{\textbf{023\_086}} & \multicolumn{2}{c}{\textbf{382\_041}} & \multicolumn{2}{c}{\textbf{640\_024}} & \multicolumn{2}{c}{\textbf{796\_006}} & \multicolumn{2}{c}{\textbf{796\_077}} & \multicolumn{2}{c}{\textbf{124\_060}} & \multicolumn{2}{c}{\textbf{147\_016}} & \multicolumn{2}{c}{\textbf{206\_013}} & \multicolumn{2}{c}{\textbf{089\_005}} & \multicolumn{2}{c}{\textbf{115\_067}} & \multicolumn{2}{c}{\textbf{574\_030}} & \multicolumn{2}{|c}{\textbf{084}} & \multicolumn{2}{c}{\textbf{090}} & \multicolumn{2}{c}{\textbf{546}} & \multicolumn{2}{|c}{\textbf{Total}} \\
  % \textbf{Q} & \multicolumn{2}{c}{\textbf{Q1}} & \multicolumn{2}{c}{\textbf{Q2}} & \multicolumn{2}{c}{\textbf{Q3}} & \multicolumn{2}{c}{\textbf{Q4}} & \multicolumn{2}{c}{\textbf{Q5}} & \multicolumn{2}{c}{\textbf{Q6}} & \multicolumn{2}{c}{\textbf{Q7}} & \multicolumn{2}{c}{\textbf{Q8}} & \multicolumn{2}{c}{\textbf{Q9}} & \multicolumn{2}{c}{\textbf{Q10}} & \multicolumn{2}{c}{\textbf{Q11}} & \multicolumn{2}{c}{\textbf{Q12}} &  \multicolumn{2}{c}{\textbf{Q13}} & \multicolumn{2}{c}{\textbf{Q14}} & \multicolumn{2}{c}{\textbf{Q15}} & \multicolumn{2}{|c}{\textbf{Total}} \\

  \textbf{Question} & \textbf{OH} & \textbf{RQ} & \textbf{OH} & \textbf{RQ} & \textbf{OH} & \textbf{RQ} & \textbf{OH} & \textbf{RQ} & \textbf{OH} & \textbf{RQ} & \textbf{OH} & \textbf{RQ} & \textbf{OH} & \textbf{RQ}  & \textbf{OH} & \textbf{RQ}  & \textbf{OH} & \textbf{RQ}  & \textbf{OH} & \textbf{RQ}  & \textbf{OH} & \textbf{RQ}  & \textbf{OH} & \textbf{RQ}  & \textbf{OH} & \textbf{RQ}  & \textbf{OH} & \textbf{RQ}  & \textbf{OH} & \textbf{RQ}  & \textbf{OH} & \textbf{RQ} \\
\midrule
GScream &   0 & 0 & 1 & 0 & 1 & 6 & 0 & 0 & 3 & 3 & 0 & 1 & 0 & 1 & 3 & 3 & 0 & 0 & 0 & 0 & 0 & 0 & 7 & 13 & 0 & 0 & 1 & 4 & 0 & 0 & 20 & 31        \\
Infusion &  5 & 5 & 5 & 7 & 6 & 4 & 0 & 0 & 4 & 5 & 4 & 17 & 0 & 5 & 9 & 7 & 15 & 16 & 0 & 0 & 3 & 1 & 7 & 1 & 1 & 1 & 1 & 5 & 0 & 0 & 61 & 75 \\
\midrule
Ours & \textbf{38} & \textbf{38} & \textbf{37} & \textbf{36} & \textbf{36} & \textbf{33} & \textbf{43} & \textbf{43} & \textbf{36} & \textbf{35} & \textbf{39} & \textbf{25} & \textbf{43} & \textbf{37} & \textbf{31} & \textbf{33} & \textbf{28} & \textbf{27} & \textbf{43} & \textbf{43} & \textbf{40} & \textbf{42} & \textbf{29} & \textbf{29} & \textbf{42} & \textbf{42} & \textbf{41} & \textbf{34} & \textbf{43} & \textbf{43} & \textbf{564} & \textbf{539}   \\
\bottomrule
\bottomrule
\end{tabular}
}

\label{tab:user_full}
\end{table*}

\begin{table}[htbp]
    \centering
    \caption{Quantitative comparison with video inpainting methods~\cite{zhou2023propainter, liang2025driveeditor} on the Waymo Open Dataset~\cite{waymodataset}. The \textbf{best} and \underline{second-best} metrics are highlighted in \textbf{bold} and \underline{underline}, respectively. We report \textbf{Time$\downarrow$} for the average extra processing time measured in seconds.}
\fontsize{8pt}{9.6pt}\selectfont {    \begin{tabular}{l|ccc}
    \toprule
        Method &  \textbf{LPIPS$\downarrow$} & \textbf{FID$\downarrow$} & \textbf{Time$\downarrow$} \\
    \midrule
    ProPainter~\shortcite{zhou2023propainter} & \textbf{0.207} & 81.925 & 147 \\
    DriveEditor~\shortcite{liang2025driveeditor} & \underline{0.225} & \underline{78.937} &  \underline{110} \\
    \midrule
    Ours & 0.237 & \textbf{74.124} & \textbf{64} \\
    \bottomrule
    \end{tabular}}
    \label{tab:supp_exp}
\end{table}

\clearpage

\end{document}